\documentclass[]{article}
\usepackage[utf8]{inputenc}
\usepackage[super,square,numbers]{natbib}
\usepackage[colorlinks=true, linkcolor=blue, citecolor=blue, urlcolor=blue]{hyperref}
\usepackage[parfill]{parskip}
\usepackage{amssymb}
\usepackage{adjustbox}

\usepackage{mathtools}
\usepackage{amsmath}
\usepackage{bm}

\providecommand{\keywords}[1]
{
  \small	
  \textbf{\textit{Keywords---}} #1
}

\title{Point Cloud Network: An Order of Magnitude Improvement in Linear Layer Parameter Count}
\makeatletter
\renewcommand\@date{{%
  \vspace{-\baselineskip}%
  \large\centering
  \begin{tabular}{@{}c@{}}
    Charles Hetterich\textsuperscript{a} \\
    \normalsize \texttt{chetterich@utexas.edu}
  \end{tabular}%

  \today
}}
\makeatother

\begin{document}

\maketitle

\noindent
\textsuperscript{a}Master of Data Science Student, University of Texas at Austin, Austin, TX, USA

\begin{abstract}
This paper introduces the Point Cloud Network (\textbf{PCN}) architecture, a novel implementation of linear layers in deep learning networks, and provides empirical evidence to advocate for its preference over the Multilayer Perceptron (\textbf{MLP}) in linear layers. We train several models, including the original \textbf{AlexNet}, using both MLP and PCN architectures for direct comparison of linear layers (Krizhevsky et al., 2012). The key results collected are model parameter count and top-1 test accuracy over the \textbf{CIFAR-10} and \textbf{CIFAR-100} datasets (Krizhevsky, 2009). AlexNet-PCN$_{16}$, our PCN equivalent to AlexNet, achieves comparable efficacy (\textit{test accuracy}) to the original architecture with a \textbf{99.5\%} reduction of parameters in its linear layers. All training is done on cloud \textit{RTX 4090} GPUs, leveraging pytorch for model construction and training. Code is provided for anyone to reproduce the trials from this paper.
\end{abstract}


\section{Introduction}
The Multilayer Perceptron is the simplest type of Artificial Neural Network (\textbf{ANN}). Since its inception in the mid-20th century, it has held firmly as one of the most popular structures in deep learning. MLPs were the first networks used with backpropagation and are relied on heavily in the attention mechanisms of the popular transformer architectures \cite{Rumelhart1986, Vaswani2017}.

Typically, networks that employ MLPs suffer from an extremely large parameter count. This is because the amount of trainable parameters present in an MLP scale by $O(n^2)$ relative to the number of input features. A parameter count so large that models have to be run across several GPUs because the parameters alone cannot fit into just one. GPT-3 and GPT-4 are two well-known models today, both of which rely on MLPs in their transformer architectures, with GPT-3 holding 175 billion trainable parameters \cite{Brown2020, OpenAI2023}.

AlexNet, widely regarded as the catalyst of the modern deep learning boom over a decade ago, popularized the convolutional operation \cite{Krizhevsky2012}— the key feature of the convolution being its reduction in parameter count in processing image data \cite{LeCun1989}.

Despite the value demonstrated by the parameter reduction present in convolutional networks, MLPs are still prevalent simply because there is currently no accessible alternative implementation of linear layers. The PCNs presented in this paper cut the parameter count present in linear layers by \textbf{an order of magnitude}, $O(n^2) \rightarrow O(n)$, while still maintaining a comparable efficacy to their equivalent MLP counterpart.

\subsection{Related Work}
Low-rank compression of ANNs is an emerging area of research which is closely related to PCNs \cite{Eo2023}. Most work in this area relies on \textit{Singular Value Decomposition} and can be divided into one of two categories (1) finding a low-rank factorization of a pre-trained network \cite{Eo2023, Idelbayev2020}, or (2) training a low-rank network directly \cite{Kamalakara2022, Vodrahalli2022}. The latter is more closely related to a PCN.

\subsection{Contribution}
This work offers a rephrasing of the same problem that low-rank factorization networks aim to solve. In low-rank factorization, we start with a weight matrix, $W$, and look to find an optimal \textit{compression} that maintains efficacy \cite{Eo2023}. The PCN starts with an already small set of parameters, and looks to find an optimal \textit{expansion} of those parameters that will perform with comparable efficacy to $W$.

We outline a light-weight implementation of the PCN architecture that is practical and generalizable to most existing deep learning networks, with source code that makes it trivial to implement.

We also provide a set of key results that demonstrate that a PCN can substantially reduce the number of parameters in linear layers while still maintaining a comparable efficacy to an MLP.

\section{Background– Multilayer Perceptron Architecture}
Two terms commonly used in describing ANNs are \textbf{neurons} and \textbf{weights}. In MLPs, neurons are the space where outputs from one layer and inputs to the next layer may be found. The weights are the \textit{things in between neurons}. They are what processes information from one layer to the next. In most current deep learning architectures this is where nearly all of the trainable parameters can be found.

Let's say we have two layers of neurons in a deep neural network, $l_i$, and $l_{i+1}$, holding $n$ and $m$ neurons, respectively. $l_i$ takes input array $x_i$ and processes that through $l_{i+1}$ into $x_{i+1}$. Between these two layers there will be trainable parameters $W_i$, a matrix of size $n \times m$. There is a also bias term, $b_{i+1}$, an array of size $m$. We define the MLP forward function as,

$$
x_{i+1} = x_i \cdot W_i + b_{i+1}
$$

noting that this operation contains $O(mn)$ trainable parameters.
\begin{figure}
  \centering
  \includegraphics[width=100mm]{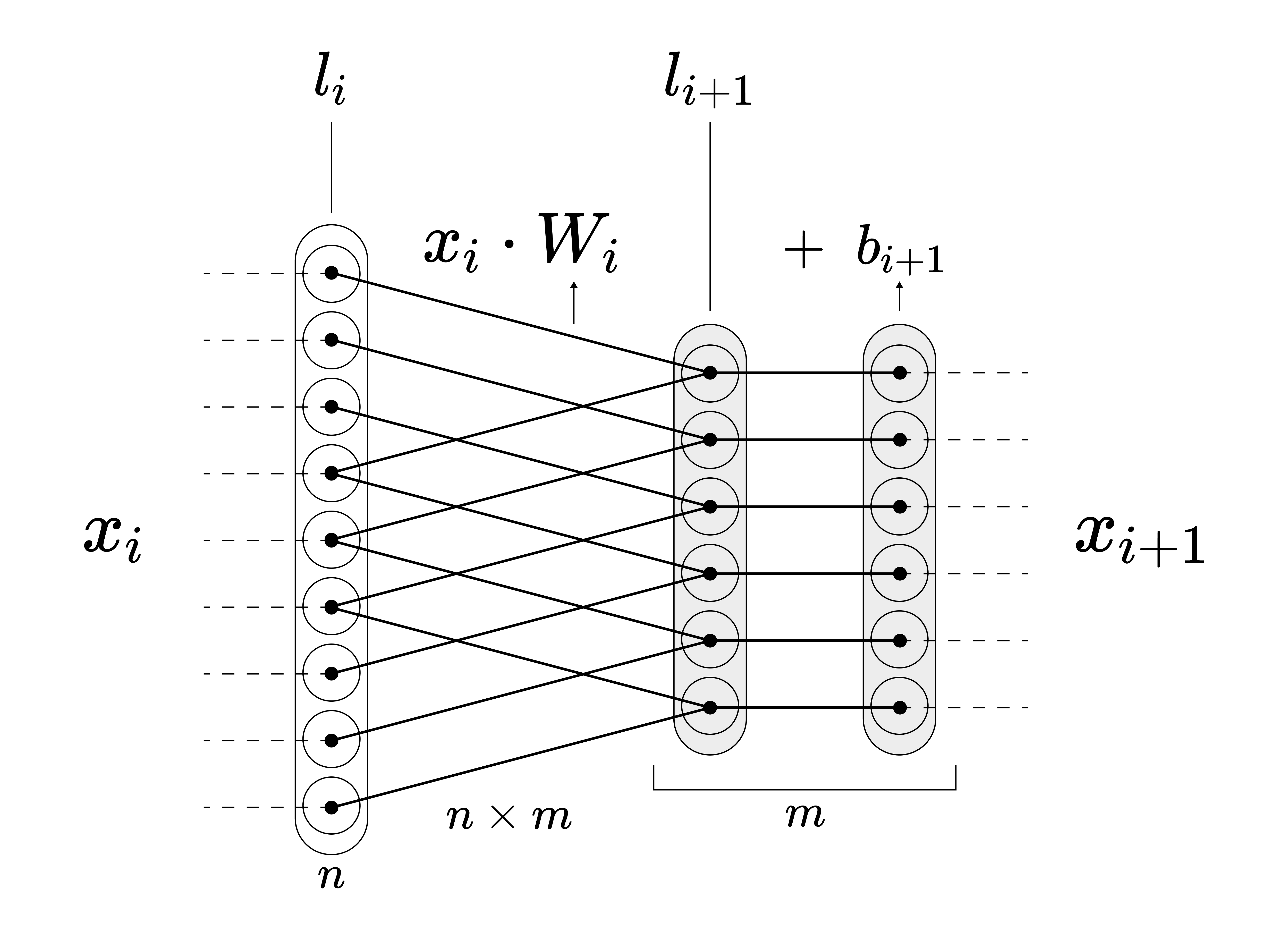}
  \caption{visual representation of MLP forward function}
  \label{fig:2.1}
\end{figure}

\section{Point Cloud Network Architecture}
In contrast to an MLP, the trainable parameters of a PCN are all \textit{neuron-centric}. What is learned are \textit{features of the neurons themselves}, rather than \textit{something in between}. In an MLP, we would say that the bias term, $b$ is \textit{neuron-centric}, but not $W$ which contains a large majority of MLP parameters.

We will treat the features of neurons as positional information (i.e. each neuron is a point in space, hence the name). The rest of this section explains step-by-step how to use these neuron features to process input data in the same way, and with the same expressiveness, as an MLP.

\subsection{Distance Matrix}
Going back to the prior example network— this time we'll say $l_i$, and $l_{i+1}$ are actually trainable parameters, where $l_i$ is of shape $n \times d$ and $l_{i+1}$ is of shape $m \times d$. $d$ is a hyperparameter representing the number of features each of our neurons have, or we can say this is the \textit{dimensionality} of the space our neurons exist in. $d$ is an especially interesting hyperparameter because it allows us to scale up or down the number of parameters in our network without affecting the number of features in a given layer. We'll also use bias term $b_{i+1}$ of size $m$ again.

The $W$ from an MLP is of shape $n \times m$. In this step, we can generate an equally shaped distance matrix $D(l_i, l_{i+1})$, where $D_{j,k}$ is the distance between neurons $l_{i,j}$ and $l_{i+1,k}$.

$$
D_{j,k}(l_i, l_{i+1}) = \sqrt{\sum_{c=1}^{d} (l_{i,j,c} - l_{i+1,k,c})^2}
$$

The intention is to replace $W$ with $D$ as follows,
$$
x_{i+1} = x_i \cdot D(l_i, l_{i+1}) + b_{i+1}
$$

However, $D$ only contains \textit{nonnegative} numbers, whereas $W \in \mathbb{R}^{n \times m}$. Using $W$, a network can choose to \textit{flip} and \textit{scale} the signal passed forward from one neuron to another, whereas using $D$, a network can only \textit{scale} signals. This would make our network using $D$ fundamentally less expressive than one using $W$. $D$ is also prone to exploding/vanishing gradients.

In this work $D$ is given as the euclidean distance between $d$-dimensional points, but $D_{j,k}$ has many possible implementations. The important feature of $D$ is that it outputs an appropriately shaped matrix that facilitates interaction between every neuron in $l_i,$ with every neuron in $l_{i+1}$. An example of an alternate implementation would be to omit the square root in the definition above. Another example would be the product $l_il_{i+1}^\top$ giving $D$ a similar property to the multiplication of \textit{keys and queries} in transformers \cite{Vaswani2017} or $UV^\top$ in low-rank factorization \cite{Idelbayev2020, Kamalakara2022, Vodrahalli2022, Eo2023}. There likely exists a more optimal definition of $D$ than the one defined here.

\subsection{Distance-Weight-Function}
The \textit{distance-weight-function}, denoted here as $F$, is an element-wise function to pass $D$ through. The goal of $F$ is to project $D$ into a space that makes it as expressive as $W$ and to provide regularization properties. In this paper the \textbf{triangle wave} is selected for $F$. Let $F_{\lambda, \epsilon}$ be an element-wise triangle wave function centered around $0$ with amplitude $\lambda$ and period $\epsilon$, with a regularization term included.

$$
F_{\lambda, \epsilon}(z) = \bm{\frac{1}{\sqrt{n}}} \cdot \frac{\lambda}{\epsilon} \cdot (\epsilon - |z \bmod{2\epsilon} - \epsilon| - \frac{\epsilon}{2})
$$

There is room for simplification, but the above equation is what is used in this work. $\bm{\frac{1}{\sqrt{n}}}$ is selected as the regularization term in order to maintain a stable signal moving forward through the network, agnostic of layer size and network depth. This term was found through a trial-and-error approach observing the variance of signal passed through untrained networks, which can be found in the provided source code. A better regularization term likely exists.

\textbf{Selection of The Triangle Wave.}
The triangle wave is selected for two desirable properties. Firstly, it takes any number $\in \mathbb{R}$ and clamps it to the range $[-\lambda, \lambda]$. This provides important control over the stability of our signal moving forward through the network, ensuring that no weights are excessively large in magnitude, regardless of how much neurons may explode away from, or implode into one another during the learning process. This in turn allows for a steady flow of gradients during backpropagation.

The second property that is specific to the triangle wave is its constant gradient and continuity— informed by the prevalence of the ReLU shape for nonlinearities \cite{Krizhevsky2012}. Cos/sin have saddle points where gradients may get stuck. Square waves' gradients are flat and saw waves are discontinuous which may lead to the network \textit{pushing} or \textit{pulling} a weight \textit{up} or \textit{down} the saw wave's drop off.

\begin{figure}
  \centering
  \includegraphics[width=100mm]{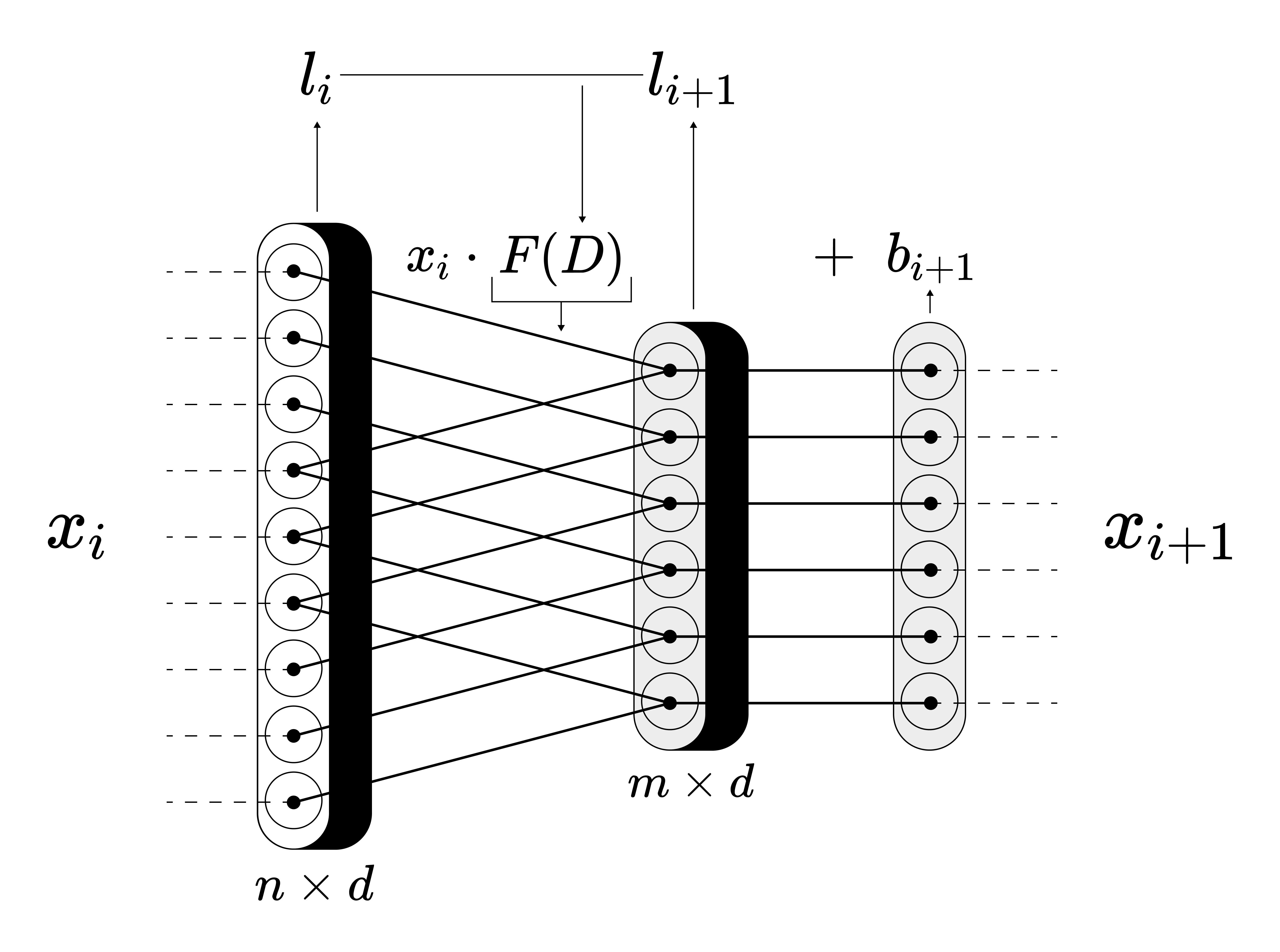}
  \caption{visual representation of PCN forward function}
  \label{fig:3.1}
\end{figure}

\subsection{Forward Function}
The PCN forward function is given as follows,

$$
x_{i+1} = x_i \cdot F_{\lambda, \epsilon}(D(l_i, l_{i+1})) + b_{i+1}
$$

which has $O(n + m)$ trainable parameters, in contrast to the $O(nm)$ trainable parameters in an MLP.

\section{Training}
This section details the model architectures implemented as well as the full training process.

Techniques such as random image augmentation, batch normalization, or residual connections are refrained from being used in favor of the direct comparison of linear layer performance of MLPs and PCNs over achieving state of the art (\textbf{SOTA}) performance. Additionally, a limited compute budget informs several design choices seen in this section.

\subsection{Model Definitions}
A modest variety of model categories are trained to evaluate the PCNs performance in different circumstances. For each model category there is a single baseline model that uses MLPs and one or more equivalent PCN models. All PCN models use hyperparameters $\lambda=1,\epsilon=0.1$.

\begin{figure}
  \centering
  \begin{adjustbox}{center}
    \includegraphics[width=120mm]{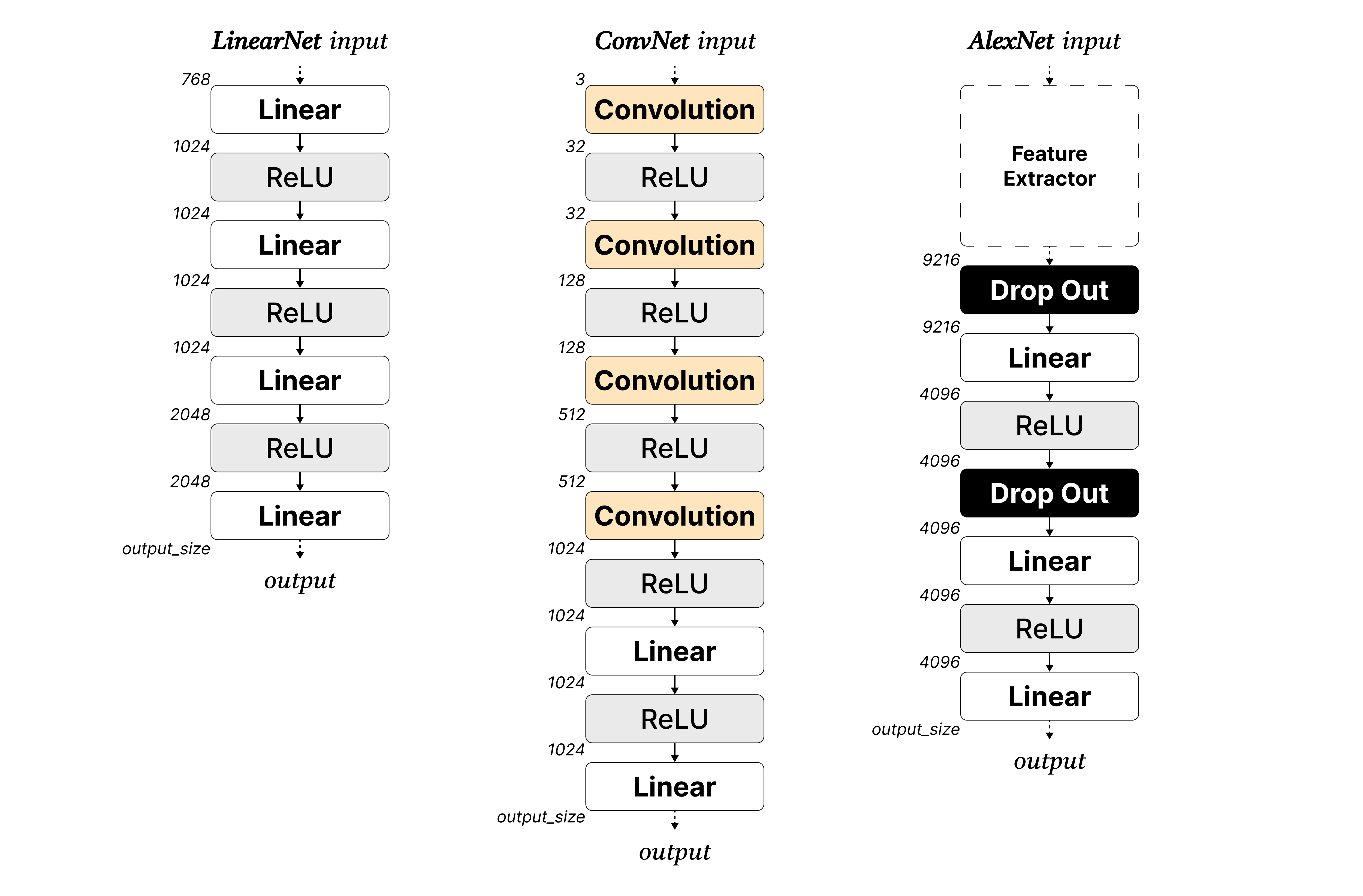}
  \end{adjustbox}
  \caption{Illustration of LinearNet, ConvNet, and AlexNet architectures. On the left side of each network are the sizes of the signal passed forward through the network.}
  \label{fig4.1}
\end{figure}

Details about the shape and depth of each network can be found in figure \ref{fig4.1}.

\subsubsection{LinearNet}
\textbf{Baseline Network.} LinearNet-MLP consists of linear layers followed by ReLUs. A linear layer with no ReLU is applied to produce the final output.

\textbf{PCN Network.} Four LinearNet-PCN$_d$ models are trained, each differing only in their dimensionality ($d \in [4, 8, 16, 32]$). Each PCN takes the baseline definition and replaces all MLP layers with equally shaped PCN layers with no further modification.

\subsubsection{ConvNet}
\textbf{Baseline Network.} ConvNet-MLP has a \textit{feature extractor} network consisting of convolutional layers followed by ReLUs. The feature extractor network is then fed into the \textit{classifier} network, consisting of linear layers followed by ReLUs with a linear layer at the end.

\textbf{PCN Network.} ConvNet-PCN$_{16}$ uses the same feature extractor as ConvNet-MLP. For the classifier a PCN is used instead of an MLP for linear layers with no further modifications.

\subsubsection{AlexNet}
\textbf{Baseline Network.} AlexNet-MLP is an untrained replica of the original model with a single modification made to the last linear layer in order to output the appropriate number of class predictions for each of CIFAR-10 and CIFAR-100. Like the previous ConvNet, AlexNet also consists of a convolutional \textit{feature extractor} network followed by a linear \textit{classifier} network. The classifier network employs \textit{dropout$_{\bm{=0.5}}$} before each linear layer, besides an isolated linear layer at the end \cite{Krizhevsky2012}.

\textbf{PCN Network.} AlexNet-PCN$_{16}$ uses the same feature extractor as AlexNet-MLP. In the classifier, MLPs are replaced with PCNs for linear layers. AlexNet-PCN$_{16}$ also features \textit{dropout$_{\bm{=0.5}}$} layers as used in the original.

\subsection{Datasets}
CIFAR-10 and CIFAR-100 are two popular image classification datasets. Both are labeled subsets of the tiny images dataset. CIFAR-10 consists of 60000 32x32 images divided into 10 classes, with 6000 images per class. The dataset is split into 50000 training images and 10000 withheld test images with exactly 1000 images of each class in the test set. CIFAR-100 is the same as CIFAR-10 but with 600 images per class, and follows the same principal for train/test split. The images and classes used in CIFAR-10 are mutually exclusive from those in CIFAR-100 \cite{Krizhevsky2009}.

The CIFAR datasets are chosen for benchmarks in order to strike balance between task difficulty and compute required. The MNIST dataset is too easy to solve— very small networks can achieve close to 100\% test accuracy— so it is difficult to extract conclusive results about a PCN's efficacy in comparison to an MLP on this dataset. ImageNet, the dataset AlexNet was originally trained on, would require too much compute. The CIFAR datasets are sufficiently difficult tasks, while also being small enough to train the largest models in a reasonable amount of time given compute constraints.

Although MNIST is not used as a benchmark in this work, it was a valuable resource in performing rapid preliminary testing of the PCN architecture. The MNIST dataset was used in making all of the architecture and regularization choices seen throughout this paper \cite{Deng2012}.

\subsection{Preprocessing}
For training/validation of LinearNet models, images are scaled down from 32x32 to 16x16, reducing the first linear layer's input size from $3072 \rightarrow 768$. Conversely, All images are scaled up to 227x227 for AlexNet models to match the original paper \cite{Krizhevsky2012}.

\subsection{Initialization}
MLP and convolutional parameters use default initializations given by \textit{torch.nn.Linear} and \textit{torch.nn.Conv2d}, respectively.

PCN neuron positional values are initialized uniformly over the range $[-1, 1]$ and bias terms uniformly over the range $[-0.1, 0.1]$.

\subsection{Loss, Gradient, and Optimizers}
Loss for all models are calculated using \textit{torch.nn.CrossEntropyLoss}, and parameter gradients are calculate using pytorch's autograd feature.

MLP and convolutional parameters are updated with \textit{stochastic gradient descent} (\textbf{SGD}), via, \textit{torch.optim.SGD}.

PCN parameters are updated with a slightly modified version of SGD that is informed by layer size. Given layer size $n$, PCN parameters $l_i$, $b_i$, gradients $\Delta l_i$, $\Delta b_i$, and learning-rate $\gamma$ we perform a PCN's SGD update as follows:

$$
\begin{aligned}
&l_i \coloneqq l_i - \gamma \Delta l_i \bm{\frac{n}{\log_{2}n}} \\
&b_i \coloneqq l_i - \gamma \Delta b_i \bm{10^5}
\end{aligned}
$$

Both of the terms $\frac{n}{\log_{2}n}$ and $10^5$ are used in order to make parameters throughout the network learn at close to the same rate, agnostic of layer size. These values were selected in early tests by observing the variance in gradients during the training process over a variety of network shapes. This optimization strategy does not account for irregularities in gradients resulting from network depth and artifacts of this fact may become pronounced in the loss/accuracy curves when attempting to train deep PCN networks, although residual connections may alleviate this problem \cite{He2015}. For the purpose of the trials done in this paper, the above optimization strategy is sufficient.

\subsection{Training Loop Details}
All models are trained with a \textbf{batch size of 1024} and \textbf{learning-rate of 0.0001}, for \textbf{3.5k epochs}. For each training iteration, we aggregate the \textit{loss}, and for each epoch we aggregate both \textit{training accuracy} and \textit{test accuracy}, seen in figure \ref{fig:a.1}.

\section{Results}
This section presents the results of training all LinearNet, ConvNet, and AlexNet architectures over the CIFAR-10 and CIFAR-100 datasets. Key results are collected in table \ref{tab:1}. Reported train/test accuracies are generated with the the final models after training. During the training of ANNs, it is normal for accuracies to fluctuate from epoch-to-epoch which introduces minor variance into these results. Loss, training accuracy, and test accuracy curves are collected in figure \ref{fig:a.1} of the appendix, which display more stable trends.

Discussion focuses on linear parameter count and test accuracy.

\begin{table}[t]
\begin{center}
\small
\begin{tabular}{ p{2.4cm}||p{2.4cm}|p{1.3cm}|p{1.5cm}|p{1.3cm}|p{1.5cm}}
\multicolumn{2}{c}{} & \multicolumn{2}{|c|}{CIFAR-10} & \multicolumn{2}{c}{CIFAR-100} \\
\hline
\footnotesize model & \textbf{\footnotesize \# linear params (millions)} & \footnotesize top-1 acc. (train) & \textbf{\footnotesize top-1 acc. (test)} & \footnotesize top-1 acc. (train) & \textbf{\footnotesize top-1 acc. (test)} \\
\hline
\footnotesize LinearNet-PCN$_{4}$ & \textbf{0.024} & 48.4 & \textbf{46.0} & 23.5 & \textbf{20.9} \\
\footnotesize LinearNet-PCN$_{8}$ & \textbf{0.044} & 53.4 & \textbf{48.9} & 26.2 & \textbf{22.7} \\
\footnotesize LinearNet-PCN$_{16}$ & \textbf{0.083} & 61.4 & \textbf{53.0} & 31.9 & \textbf{26.1} \\
\footnotesize LinearNet-PCN$_{32}$ & \textbf{0.161} & 66.9 & \textbf{52.8} & 39.1 & \textbf{28.1} \\
\footnotesize LinearNet-MLP & \textbf{3.957} & 96.8 & \textbf{52.1} & 78.7 & \textbf{25.2} \\
\hline
\hline
\footnotesize ConvNet-PCN$_{16}$ & \textbf{0.035} & 88.4 & \textbf{60.0} & 72.9 & \textbf{26.1} \\
\footnotesize ConvNet-MLP & \textbf{1.06} & 98.9 & \textbf{58.1} & 99.8 & \textbf{27.3} \\
\hline
\hline
\footnotesize AlexNet-PCN$_{16}$ & \textbf{0.296} & 85.7 & \textbf{78.9} & 51.6 & \textbf{43.7} \\
\footnotesize AlexNet-MLP & \textbf{54.575} & 84.1 & \textbf{78.6} & 52.5 & \textbf{47.5} \\
\hline
\end{tabular}
\end{center}
\vspace{-.5em}
\caption{Train and test accuracies (\%) over both CIFAR-10 and CIFAR-100 datasets for each model, along with the parameter counts of their linear layers.}
\label{tab:1}
\vspace{-.5em}
\end{table}

\subsection{LinearNet}
Four LinearNet-PCN$_d$ models and LinearNet-MLP are trained. For $d = 4,8$ there is a degradation in performance relative to the MLP. At $d = 16,32$, the PCN outperforms the MLP on both datasets. As $d$ increases, the PCNs experience more overfitting. The MLP experiences substantially more overfitting than all PCNs. LinearNet-PCN$_{32}$, the largest PCN in this class of models, has 161k parameters, which is a \textbf{95.9\%} reduction from 3.95 million parameters in the MLP. Additionally, figure \ref{fig:a.1} displays a consistent increase in PCN performance with an increase in $d$.

\subsection{ConvNet}
Both ConvNet-MLP and ConvNet-PCN$_{16}$ have 5.35 million convolutional parameters. ConvNet-PCN$_{16}$ outperforms ConvNet-MLP by \textbf{1.9\%} on CIFAR-10 and underperforms by \textbf{1.2\%} on CIFAR-100. Similarly to LinearNet, ConvNet-MLP experiences more overfitting than ConvNet-PCN$_{16}$. ConvNet-PCN$_{16}$ has 35k linear parameters, which is a \textbf{96.7\%} reduction from 1.06 million linear parameters in the MLP.

\subsection{AlexNet}
Both AlexNet-MLP and AlexNet-PCN$_{16}$ have 2.47 million convolutional parameters. AlexNet-PCN$_{16}$ outperforms AlexNet-MLP by \textbf{0.3\%} in CIFAR-10, and underperforms by \textbf{3.8\%} in CIFAR-100. Both models experience similar amounts of overfitting. AlexNet-PCN$_{16}$ has 296k linear parameters, which is a \textbf{99.5\%} reduction from 54.6 million linear parameters in the MLP.

\section{Limitations and Future Work}
\subsection{Memory Requirements}
As has been demonstrated by this work, the PCN architecture can substantially reduce the number of parameters needed to train linear layers. However, the implementation seen here does not actually reduce the memory requirements. This is due to my reliance on pytorch's native autograd feature and \textit{torch.cdist} to find $D$. During the forward pass, $D$ in its entirety is calculated and stored in memory, which is the same size as $W$. A fused kernel function for calculating $x_{i+1,k}$ that never stores $D$ but instead calculates $D_{j,k}$ as needed could be used.

$$
\sigma_k(x_i) = b_{i+1,k} + \sum_{j=1}^{n} x_{i,j} \cdot F_{\lambda, \epsilon}(D_{j,k}(l_{i,j}, l_{i+1,k}))
$$

Successfully implementing this along with its corresponding gradient functions on accelerated hardware would reduce memory consumption $O(n^2) \rightarrow O(n)$ during training and inference.

\subsection{Compute Requirements}
Two limiting factors of deep learning are \textbf{memory} and \textbf{compute}. The PCN architecture can alleviate memory consumption, but requires $O(d)$ times more compute than an MLP.

\subsection{Network Stability}
As has been stated previously, all regularization terms used in the PCNs presented in this work were found through trial-and-error rather than rigorous math. Because of this, these PCNs are not resilient to their hyperparameters and a more robust PCN definition should be investigated.

\subsection{Applying PCNs Elsewhere}
In this work the PCN architecture is applied to linear layers. The same concept can be applied to the convolutional layers along the \textit{channel} axis, and to graph layers along the \textit{node-feature} axis.

\begin{figure}[h]
  \centering
  \includegraphics[width=120mm]{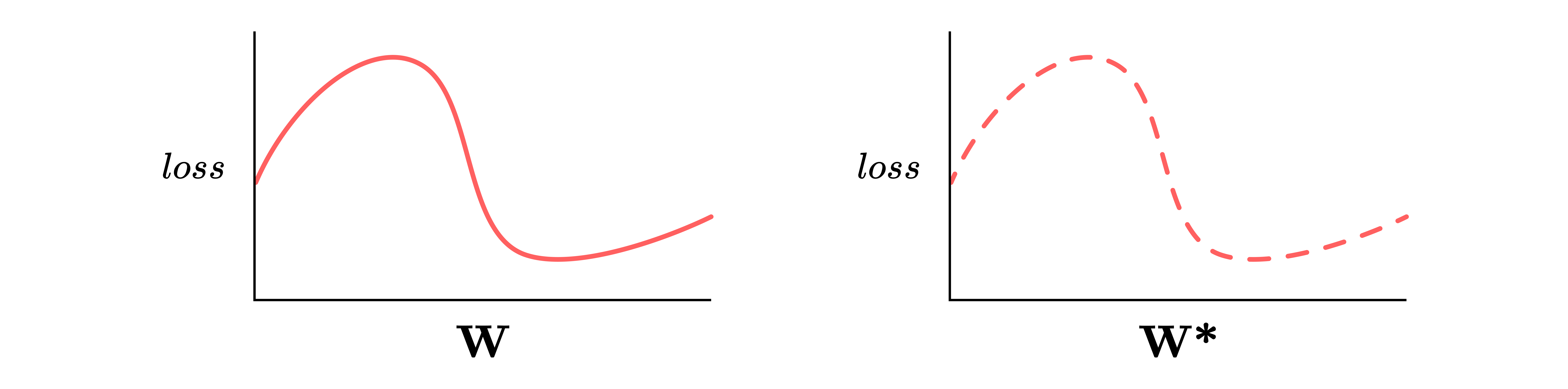}
  \caption{On the left: loss distributed across $\bm{W}$. On the right: loss distributed across $\bm{W^*}$, \textit{projected into the space of} $\bm{W}$. Blank space represents where $\bm{W^*}$ does not occur.}
  \label{fig7.1}
\end{figure}
\subsection{Conjecture— Why PCNs Work}
 
The concept of a PCN can be boiled down to an MLP where we generate a plausible $W$, similarly to low-rank factorization \cite{Idelbayev2020, Kamalakara2022, Vodrahalli2022, Eo2023}. Let $\bm{W} = \mathbb{R}^{n \times m}$ be the set of all possible values for $W$, $\bm{W^*} \subseteq \bm{W}$ be the set of all possible values for $F(D)$, and $\overline{L}_{\bm{W}}$ be the mean loss w.r.t. $\bm{W}$.  If $\overline{L}_{\bm{W^*}}=\overline{L}_{\bm{W}}$, then $F(D)$ should have a comparable efficacy to $W$. Consequently, if $\overline{L}_{\bm{W^*}} < \overline{L}_{\bm{W}}$ or $\overline{L}_{\bm{W^*}} > \overline{L}_{\bm{W}}$, then $F(D)$ would be expected to perform better or worse than $W$, respectively.

It may be interesting to investigate $F(D)$ that \textit{maximizes} $\overline{L}_{\bm{W}} - \overline{L}_{\bm{W^*}}$.

\section{Ethical Concerns}
With the exception of recent high profile publications, it seems a relatively uncommon practice to include an ethics section in a deep learning paper like this one. I use this section as a platform to attempt to mindfully outline some of my concerns. I include this section to advocate for a culture within academia that normalizes, legitimizes, and prioritizes this conversation— hoping that a more organized practice forms.

\textbf{Downstream Consequences.} Deep Learning is a unique technology in that it is largely task-agnostic. Because of this, the set of downstream applications is uncharacteristically large compared to other technology. Although the PCNs presented in this paper are applied to test datasets, the intention is to integrate this into existing deep learning architectures for which there are existing harmful applications. This makes it important to be cognizant of and acknowledge these harmful applications.

\textbf{Mindful Conversations.} Having productive conversations about A.I. safety is a bit paradoxical. It is surely helpful to be aware of potential negative applications of deep learning, yet it may actually be harmful to indulge in any unnecessary details that don't move the conversation forward. For example, I would consider media outlets echoing unproductive details about harmful applications to be an unethical practice.

\section{Acknowledgements}
I would like to thank Ryan Schaake for offering fruitful comments, review, and insight.

\bibliographystyle{plainnat}

\newpage
\appendix
\section{Supplemental Material}
\subsection{Source Code}
Source code and other materials can be found at \url{https://gitlab.com/cHetterich/pcn-paper-and-materials}.

\subsection{Figures}
\begin{figure}[h]
  \centering
  \includegraphics[width=120mm]{fig_a.1.png}
  \caption{A collection of training accuracy, test accuracy, and loss curves for all models.}
  \label{fig:a.1}
\end{figure}

\end{document}